\documentclass[10pt,twocolumn,letterpaper]{article}

\usepackage[pagenumbers]{cvpr} 

\usepackage{graphicx}
\usepackage{amsmath}
\usepackage{amssymb}
\usepackage{booktabs}

\usepackage[pagebackref,breaklinks,colorlinks]{hyperref}

\usepackage[capitalize]{cleveref}
\crefname{section}{Sec.}{Secs.}
\Crefname{section}{Section}{Sections}
\Crefname{table}{Table}{Tables}
\crefname{table}{Tab.}{Tabs.}

\def\cvprPaperID{*****} 
\def\confName{CVPR}
\def\confYear{2023}

\begin{document}

\title{3rd Place Solution for MOSE Track in CVPR 2024 PVUW workshop: Complex Video Object Segmentation}

\author{
    Xinyu Liu$^{1}$\qquad Jing Zhang$^{1}$\qquad Kexin Zhang$^{1}$\qquad Yuting Yang$^{1}$ \\
    Licheng Jiao$^{1}$\qquad Shuyuan Yang$^{1}$\qquad\vspace{3mm} \\
    $^{1}$Intelligent Perception and Image Understanding Lab, Xidian University
}
\maketitle

\begin{abstract}
Video Object Segmentation (VOS) is a vital task in computer vision, focusing on distinguishing foreground objects from the background across video frames. Our work draws inspiration from the Cutie model, and we investigate the effects of object memory, the total number of memory frames, and input resolution on segmentation performance. This report validates the effectiveness of our inference method on the coMplex video Object SEgmentation (MOSE) dataset, which features complex occlusions. Our experimental results demonstrate that our approach achieves a J\&F score of 0.8139 on the test set, securing the third position in the final ranking. These findings highlight the robustness and accuracy of our method in handling challenging VOS scenarios.
\end{abstract}

\section{Introduction}
\label{sec:intro}
The Pixel-level Video Understanding in the Wild (PVUW) Challenge is a computer vision competition that encompasses four different tracks. Its overarching aim is to advance the field of pixel-level scene understanding, which involves classifying object categories, generating masks, and assigning semantic labels to every pixel within an image. Given that the real world is dynamic and not limited to static images, the ability to segment and understand videos is of greater relevance and practicality for real-world applications. The PVUW Challenge is designed to spur the development of technologies that can accurately interpret and segment video content in natural settings, thereby pushing the boundaries of computer vision in dynamic scene understanding.

In the 3rd of the PVUW challenge, We have noted the addition of two new tracks in the third PVUW challenge: The MOSE Track includes additional videos and annotations that feature challenging elements such as the disappearance and reappearance of objects, inconspicuous small objects, heavy occlusions, and crowded environments\cite{ding2023mose}. The Motion Expression guided Video Segmentation (MeViS) Track is designed to advance the study of natural language-guided video understanding in complex environments, with the goal of fostering the development of a more comprehensive and robust pixel-level understanding of video scenes in such settings and realistic scenarios through the inclusion of new videos, sentences, and annotations\cite{MeViS}.

\begin{figure}[t]
  \centering
   \includegraphics[width=1\linewidth]{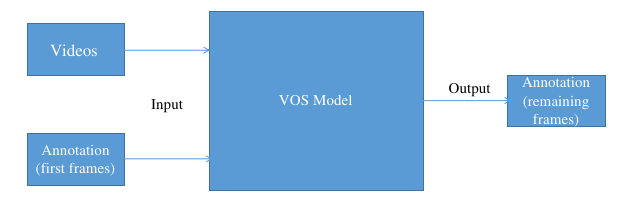}
   \caption{\textbf{VOS framework overview.} It consists of three independent components: a segmenter, a referring tracker, and a temporal refiner.
   }
   \label{fig:framework}
\end{figure}

In the realm of Video Object Segmentation (VOS), especially within the semisupervised paradigm, the objective is to track and segment objects from a broad range of categories based solely on an initial frame's annotation. VOS methodologies find extensive application in fields such as robotics\cite{petik2022learning} , video editing\cite{cheng2021modular} , and in reducing the burden of data annotation\cite{athar2023burst}. Given the Video Object Segmentation (VOS) dataset, comprising training and test sets, each frame of the video contains corresponding annotation data. These annotations are represented as two-dimensional matrices of height and width. For the pixel-wise annotation matrix, each element records the pixel information (e.g., RGB channels) of the corresponding pixel in the video frame. For the classification result matrix, each element represents a one-hot vector of length equal to the number of object categories in the VOS task, indicating the classification of the corresponding pixel (see \cref{fig:framework}).

Recent VOS methods utilize a memory-based paradigm \cite{oh2019video,cheng2022xmem,yang2021associating,bekuzarov2023xmem++}, where a memory representation is built from previously segmented frames (either provided or generated by the model). New query frames then access this memory to retrieve features for segmentation. These methods predominantly employ pixel-level matching for memory reading, whether through a single \cite{oh2019video} or multiple matching layers \cite{yang2021associating}, and construct segmentation from the pixel memory readout. Pixel-level matching independently maps each query pixel to a linear combination of memory pixels (e.g., using an attention layer). However, this approach often lacks high-level consistency and is vulnerable to matching noise, especially when distractors are present.

The complex video object segmentation task focuses on the tracking and segmentation of objects within intricate scenes. MOSE is specifically designed to test VOS models in complex environments featuring multiple objects, frequent occlusions, and numerous distractors. This dataset poses a significant challenge due to its high variability and realistic scenes, making it a more rigorous benchmark than traditional datasets like DAVIS-2017 \cite{perazzi2016benchmark}. In fact, recent methods \cite{cheng2022xmem,yang2021associating} show a performance drop of over 20 points in J \& F when evaluated on MOSE compared to DAVIS-2017.This significant performance decline on MOSE highlights the limitations of current memory-based VOS techniques. In scenarios with high object density and frequent interactions, pixel-level matching struggles with segmentation accuracy due to noise and distractors. These issues emphasize the need for more robust and context-aware approaches in memory-based VOS to effectively manage the complexities of datasets like MOSE.

Recently, the Cutie approach \cite{cheng2023putting} has restructured the video object segmentation task into three distinct subprocesses: image object segmentation, tracking/alignment, and refinement. Furthermore, Cutie construct a compact object memory to summarize object features in the long term, which are retrieved as targetspecific object-level representations during querying, showcasing significant advancements in the field of Video Object Segmentation (VOS). Leveraging the exceptional capabilities of Cutie, our team secured the 3rd position in the complex video object segmentation  track of the 3rd PVUW Challenge at CVPR 2024,  and all without the need for additional training.

\section{Method}
Our approach is inspired by recent work on video object segmentation, particularly the Cutie framework by Cheng et al. \cite{cheng2023putting}. Cutie operates in a semi-supervised video object segmentation (VOS) setting, where it takes a first-frame segmentation as input and processes subsequent frames sequentially.

Cutie encodes segmented frames into a high-resolution pixel memory \(F\) and a high-level object memory \(S\). These memories are used for segmenting future frames. When segmenting a new frame, Cutie first retrieves an initial pixel readout \(R_0\) from the pixel memory using the encoded query features. This initial readout is typically noisy due to low-level pixel matching.

To enhance this initial readout, Cutie enriches \(R_0\) with object-level semantics using information from the object memory \(S\) and object queries \(X\). This is done through an object transformer with multiple transformer blocks. The final enriched output, \(R_L\), is then passed to the decoder to generate the output mask.

In summary, Cutie introduces three main contributions: object-transformer, sec:masked-attention, and object-memory.

\begin{figure*}[t]
  \centering
   \includegraphics[width=1\linewidth]{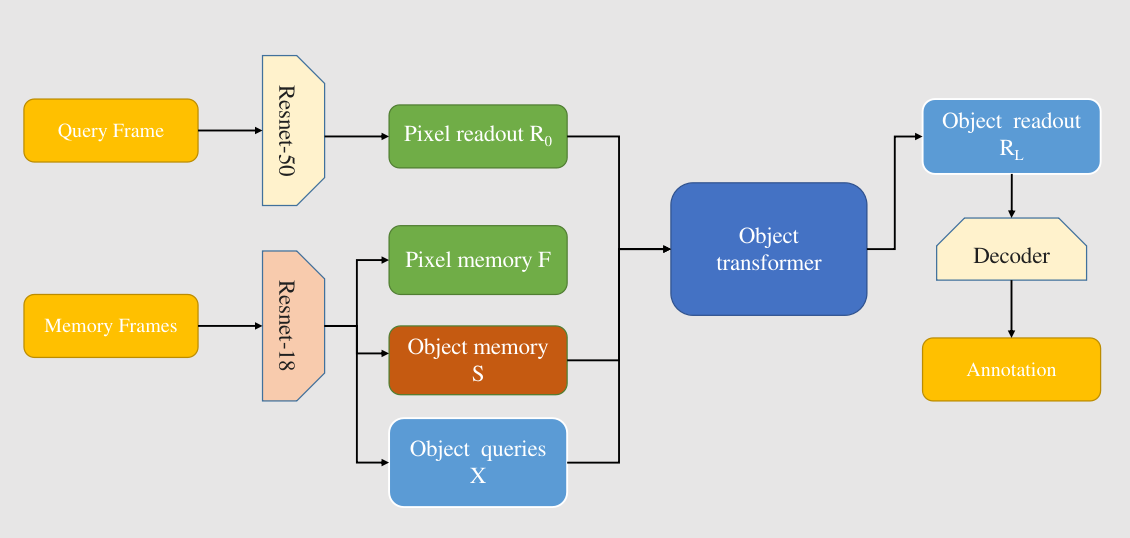}
   \caption{\textbf{The framework of Cutie.} 
   }
   \label{fig:retrack}
\end{figure*}


The 'Cutie-base' model is based on the 'base' variant, utilizing ResNet-50 as the query encoder backbone. It consists of $C=256$ channels, $L=3$ object transformer blocks, and $N=16$ object queries.

The query and mask encoders are designed using ResNets~\cite{he2016deepResNet}. Following previous studies~\cite{oh2019videoSTM,cheng2022xmem}, we discard the final convolutional stage and employ the stride 16 feature. 

The object transformer block integrates both query FFN and pixel FFN components. The query FFN comprises a 2-layer MLP with a hidden size of $8C=2048$. Meanwhile, the pixel FFN utilizes two $3\times3$ convolutions with a reduced hidden size of $C=256$ to minimize computational overhead. The ReLU activation function is employed throughout the network.

\section{Experiment}

\subsection{Inference}

When testing, the input video is upscaled to a resolution of 720p, which provides a higher density of pixel information compared to lower resolutions such as 480p. This increased resolution enhances the detail and clarity of the video, allowing for more precise segmentation and tracking of objects. The choice of 720p as the standard testing resolution is motivated by the need to balance between the level of detail required for accurate segmentation and the computational resources required to process the video data.

In the context of the memory frame encoding, we update both the pixel memory and the object memory every $r$-th frame. The default value of $r$ is set to 3, following the same configuration used in the XMem framework~\cite{cheng2022xmem}. This interval strikes a balance between capturing the temporal evolution of the scene and maintaining computational efficiency. In the attention component of the pixel memory, we retain the keys $\mathbf{k}$ and values $\mathbf{v}$ from the first frame, which is provided by the user. This initial frame serves as a reference point for the video sequence. For subsequent memory frames, we employ a First-In-First-Out (FIFO) strategy, which ensures that the most recent information is retained while older data is gradually phased out. This approach is designed to keep the memory footprint manageable and focused on the most relevant frames.The choice of a predefined limit of $T_{\max} = 15$ for the total number of memory frames is a practical compromise. This value balances the need to avoid excessive memory usage and maintain real-time performance while still capturing sufficient temporal evolution of the scene. Maintaining a history of 15 frames is generally adequate for effectively exploiting temporal correlations in VOS tasks. This enhances segmentation accuracy by providing enough context for object tracking and appearance prediction without imposing excessive computational overhead or compromising system responsiveness. Extending this limit further could lead to diminishing returns, as the additional frames may not significantly improve performance and could increase computational load unnecessarily.

The noise in the memory increases with its size, and when the sequence is long, performance can degrade. Based on these observations, we propose filtering affinities to retain only the top-$k$ entries. This effectively eliminates noise regardless of the sequence length. The top-k strategy not only enhances robustness but also overcomes the overhead associated with top-$k$ operations. The graph reports the performance improvement and robustness brought about by top-$k$ filtering. In our implementation, the top-k operation uses query filtering to refine the memory. To further manage the memory capacity, we apply top-$k$ filtering~\cite{cheng2021mivos} with $k = 60$ to the pixel memory. Setting top-$k$ to 60 has the effect of prioritizing the most relevant pixel memories based on their attention scores, which is crucial for maintaining accurate segmentation over time while preventing the memory from being overwhelmed with less significant information. This approach ensures that the memory retains the most informative aspects of the scene, which are necessary for consistent and reliable segmentation results, especially in lengthy video sequences. This filtering technique selects the top $k$ most relevant pixel memories, based on their attention scores, for updating the memory. By doing so, we prioritize the most informative pixel data, which is crucial for maintaining accurate segmentation over time, while also preventing the memory from being overwhelmed with less significant information.

In the final testing phase, we employed Test-Time Augmentation (TTA), which is a strategy that enhances the robustness and accuracy of predictions by incorporating a variety of augmented versions of the input data. TTA is a powerful technique that can help to mitigate overfitting and improve generalization by simulating variations in the data that the model might encounter in real-world scenarios. The core idea behind TTA is to make predictions not just on the original test samples but also on perturbed versions of them. These perturbations can include a range of augmentations such as scaling, cropping, flipping, and rotating the input images or videos. By applying these augmentations, we can capture a broader range of possible transformations that the objects of interest might undergo, thereby improving the model’s ability to recognize and segment them accurately.In the context of video segmentation, TTA can be particularly beneficial due to the dynamic nature of video data. Frames in a video can have significant variations due to camera motion, object motion, lighting changes, and other environmental factors. TTA helps to address these variations by providing the model with multiple perspectives of the same scene, which can lead to more consistent and reliable segmentations over time.

\section{Evaluation Metrics}
To evaluate the performance of our model, we compute the Jaccard value (J), the F-Measure (F), and the mean of J and F.

\subsection{Jaccard Value (J).}
The Jaccard value, also known as Intersection over Union (IoU), measures the similarity between two sets. For a predicted segmentation mask $P$ and a ground truth segmentation mask $G$, the Jaccard value is defined as:
\begin{equation}
    J = \frac{|P \cap G|}{|P \cup G|} = \frac{\sum_{i} P_i \cdot G_i}{\sum_{i} P_i + \sum_{i} G_i - \sum_{i} P_i \cdot G_i},
\end{equation}
where $P_i$ and $G_i$ denote the value of the $i$-th pixel in the predicted and ground truth masks, respectively. The Jaccard value ranges from $0$ to $1$, with higher values indicating better performance.

\subsection{F-Measure (F).}
The F-Measure is a metric that combines Precision and Recall, commonly used to evaluate the performance of binary classification models. It is calculated as follows:
\begin{equation}
    F = \frac{2 \cdot \text{Precision} \cdot \text{Recall}}{\text{Precision} + \text{Recall}},
\end{equation}
where
\begin{equation}
    \text{Precision} = \frac{|P \cap G|}{|P|} = \frac{\sum_{i} P_i \cdot G_i}{\sum_{i} P_i},
\end{equation}
and
\begin{equation}
    \text{Recall} = \frac{|P \cap G|}{|G|} = \frac{\sum_{i} P_i \cdot G_i}{\sum_{i} G_i}.
\end{equation}
The F-Measure also ranges from $0$ to $1$, with higher values indicating better model performance in handling positive and negative samples.

\subsection{Mean of J and F.}
To comprehensively evaluate the model's performance, we compute the mean of the Jaccard value (J) and the F-Measure (F):
\begin{equation}
    \text{Mean(J, F)} = \frac{J + F}{2}.
\end{equation}

These metrics together provide a robust assessment of the segmentation model's accuracy and consistency, offering insights into its performance in predicting segmentation masks.

\begin{table}[t]
\setlength{\tabcolsep}{3mm}
\centering
\begin{tabular}{l|ccc}
    \toprule
    User & J & F & J\&F \\
    \midrule
    PCL\_MDS & 0.7235 (1) & 0.8044 (1) & 0.7640 (1) \\
    yahooo & 0.7101 (2) & 0.7899 (3) & 0.7500 (3) \\
    valgab & 0.7095 (3) & 0.7923 (2) & 0.7509 (2) \\
    xzs123456 & 0.7011 (4) & 0.7779 (4) & 0.7395 (4) \\
    ISS & 0.6892 (5) & 0.7705 (5) & 0.7299 (5) \\
    \bottomrule
\end{tabular}
\caption{\textbf{Leaderboard during the development phase.}}
\label{tab:develop}
\end{table}

\begin{table}[t]
\setlength{\tabcolsep}{3mm}
\centering
\begin{tabular}{l|ccc}
    \toprule
    User & J & F & J\&F \\
    \midrule
    PCL\_MDS & 0.8101 (1) & 0.8789 (1) & 0.8445 (1) \\
    Yao\_Xu\_MTLab & 0.8007 (2) & 0.8683 (2) & 0.8345 (2) \\
    ISS & 0.7879 (3) & 0.8559 (3) & 0.8219 (3) \\
    xsong2023 & 0.7873 (4) & 0.8544 (4) & 0.8208 (4) \\
    yangdonghan50 & 0.7799 (5) & 0.8480 (5) & 0.8139 (5) \\
    \bottomrule
 \end{tabular}
\caption{\textbf{Leaderboard during the test phase.}}
 \label{tab:test}
\end{table}

\section{Comparison with Other Methods}
In the 1st Complex Video Object Segmentation Challenge, our(ISS) method  demonstrated significant performance improvements in both the development and test phases. The leaderboards for the development and test phases are presented in Tables \ref{tab:develop} and \ref{tab:test}, respectively. Our method achieved Jaccard values (J) and F-Measures (F) that outperformed most other participants. Specifically, in the development phase, our method attained a Jaccard value of 0.6892 and an F-Measure of 0.7705, resulting in a combined J\&F score of 0.7299. Similarly, in the test phase, our method achieved a Jaccard value of 0.7799, an F-Measure of 0.8480, and a combined J\&F score of 0.8139. These results highlight the effectiveness and robustness of our method.

\section{Conclusion}
In this study, we developed a method for Video Object Segmentation (VOS) drawing inspiration from the Cutie model. We examined key factors such as object memory management, the number of memory frames, and input resolution, assessing their impact on segmentation performance. Our approach was rigorously evaluated on the MOSE dataset, achieving a notable J\&F score of 0.8139, which earned us third place in the test phase. These results underscore the robustness of our method in addressing the challenges posed by complex occlusions in video sequences.

{\clearpage
\small
\bibliographystyle{ieee_fullname}
\bibliography{egbib}
}

\end{document}